%% file: acl_latex.tex
\pdfoutput=1
\documentclass[11pt]{article}

\usepackage{acl}

\usepackage{times}
\usepackage{latexsym}

\usepackage[T1]{fontenc}
\usepackage[utf8]{inputenc}

\usepackage{microtype}

\usepackage{xcolor}
\usepackage{colortbl}
\usepackage{hhline}
\usepackage{amsmath}
\usepackage{amssymb}
\usepackage{multirow}
\usepackage{booktabs}
\usepackage{graphicx}
\usepackage{enumitem}
\usepackage{tikz}
\usepackage{booktabs}
\usepackage{xspace}
\usepackage{centernot}
\usepackage{enumitem} % defines [nosep] for itemize

\usepackage[ruled,noline,linesnumbered]{algorithm2e}
\usepackage{algorithm2e}
\usepackage{algorithmic}
\usepackage[size=scriptsize]{todonotes}

\newcommand{\measure}{Factual Ablation\xspace}

\newcommand{\promptTask}{content transfer\xspace}

\title{Probing Factually Grounded Content Transfer with Factual Ablation}

\author{Peter West\textsuperscript{$\dagger$} \hspace{.1cm} \textbf{Chris Quirk}\textsuperscript{$\ddagger$} \hspace{.1cm} \textbf{Michel Galley}\textsuperscript{$\ddagger$} \hspace{.1cm}  \textbf{Yejin Choi \textsuperscript{$\dagger$$\star$}}\\
  \textsuperscript{$\dagger$}Paul G. Allen School of Computer Science \& Engineering, University of Washington\\
  \textsuperscript{$\ddagger$} Microsoft Research, Redmond, WA, USA
  \\
  \textsuperscript{$\star$}Allen Institute for Artificial Intelligence\\
  \texttt{\{pawest;yejin\}\@cs.washington.edu} \hspace{.1cm} \texttt{\{chrisq;mgalley\}@microsoft.com}}

\begin{document}
\maketitle
\begin{abstract}
Despite recent success, large neural models often generate factually incorrect text. Compounding this is the lack of a standard automatic evaluation for factuality--it cannot be meaningfully improved if it cannot be measured. \textit{Grounded generation} promises a path to solving both of these problems: models draw on a reliable external document (\textit{grounding}) for factual information, simplifying the challenge of factuality. Measuring factuality is also simplified--to \textit{factual consistency}, testing whether the generation agrees with the grounding, rather than all facts. Yet, without a standard automatic metric for factual consistency, factually grounded generation remains an open problem. 

We study this problem for \textit{content transfer}, in which generations extend a prompt, using information from factual grounding. Particularly, this domain allows us to introduce the notion of \textit{factual ablation} for automatically measuring factual consistency: this captures the intuition that the model should be less likely to produce an output given a less relevant grounding document. In practice, we measure this by presenting a model with two grounding documents, and the model should prefer to use the more factually relevant one. We contribute two evaluation sets to measure this. Applying our new evaluation, we propose multiple novel methods improving over strong baselines.
\end{abstract}

\input{sections/intro}
\input{sections/background}

\input{sections/methodology}
\input{sections/experiments}

\input{sections/results}
\input{sections/conclusion}

\section*{Acknowledgments}
We thank Felix Faltings and Gerold Hintz for technical and intellectual support. This work was funded in part by the Natural Sciences and Engineering Research Council of Canada (NSERC) (funding reference number 401233309), DARPA MCS program through NIWC Pacific (N66001-19-2-4031), and the Allen Institute for AI.

\bibliographystyle{acl_natbib}
\bibliography{anthology,custom}

\clearpage

\appendix

\input{sections/appendix}

\end{document}

%% file: sections/intro.tex
\section{Introduction}

\input{figs/fig_1}

Large pretrained models have shown impressive effectiveness at longstanding tasks and benchmarks. %(CITE,CITE). 
% MG: Since you are citing examples below, maybe the above cite's can go.
One exciting example is GPT-3 \citep{Brown2020LanguageMA}, which completes tasks with remarkable clarity and knowledge---without supervision---simply by writing what might come next. Yet significant challenges prevent these models from helping humans write real documents.
% Merge paragraph as next sent is in support of prev.
For example, in Figure~\ref{fig:fig_1} GPT-3 attempts to auto-complete the next sentence of a prompt regarding auto racer Ralph De Palma; GPT-3 suggests the 500-mile Indy-500 race had an impressive--yet impossible--finishing distance of ``more than 500 miles.''

%\pw{FROM YEJIN: less discussion of summarization. Focus more on content transfer, then eventually compare to summarization mentioning that FA is more natural in this setting}
Such factual hallucinations limit the usability of existing models \citep{maynez-etal-2020-faithfulness}. Issues are exacerbated by the black-box nature of memorized knowledge that these models draw from, which may have factual gaps or be out-of-date. This motivates explicitly controlling the information models generate with, by \textit{textual grounding}. 
%Summarization is a prime example of this: 
Summarization is a good example of this: 
all information needed for the summary comes from the source document (grounding). 
%But this desideratum is not restricted to summarization--any generation task requiring knowledge can be controlled with grounded information.
Besides assuring models draw on factual knowledge, introducing grounding simplifies the challenge of evaluating factuality. Rather than verifying generations against \textit{all facts}, the problem is reduced to testing \textit{factual consistency} with information in the grounding. However, measuring this automatically is an open problem.%Yet measuring this automatically is still an open problem.
% MG: sentence a bit weird as it's presented as a feature but with the word 'limited'. Captured?
%limited 
%\textit{factual consistency} 
%with respect 
%relative
%to the grounding document, rather than to \textit{all facts}.

In this work, we study factual consistency in the setting of Figure~\ref{fig:fig_1}: generating the next sentence with grounded information. We refer to this as \emph{\promptTask} \citep{prabhumoye2019towards,qin-etal-2019-conversing}--transferring knowledge from a source document to continue a target document. Factual consistency has largely been studied in summarization,
%due to an abundance of data,
but \promptTask introduces an exciting notion of control (the document being extended) which affects style, content, and factual selection. % It further lends itself to a unique approach for judging fac

%Central to any study of factual consistency is defining a way to measure it. In this work, we introduce \textit{factual ablation}--the intuitive notion that a well-grounded model will be less likely to produce a generation $y$ as its grounding $g$ contains less information relevant to $y$. Factual ablation has particular relevance to \promptTask. The underlying Wikipedia data used for this task allows us to construct a natural, factual ablation-based evaluation set to measure factual consistency in this setting.
%\pw{FROM YEJIN: give a more precise definition here, this one is confusing}
% \cq{I tried a rewrite here -- does it look okay?}
Central to any study of factual consistency is defining a way to measure it. In this work, we introduce \textit{factual ablation}, which asserts that an output $y$ should be more likely when grounding $g$ is more relevant.
In particular, if grounding $g$ entails $y$ but $g'$ does not, $p(y|g)$ should be greater than $p(y|g')$; the closer $g$ and $g'$, the more challenging the example.
An evaluation set for factual ablation is constructed by collecting such grounding pairs to test models with. Content transfer is particularly suited for this: due to continuous edits in the underlying Wikipedia data, there are many instances of document pairs $g,g'$ which are relevant to the same target document, but result in different continuations.
Following a similar intuition to factual ablation, we propose both training-time and inference-time approaches that measure the effect grounding has on generation, to keep models on-topic and factually consistent with grounding.
%\pw{Following similar intuitions to factual ablation (the signal of reduced grounding) we also propose techniques for improving factual consistency and groundedness of models. }
%\pw{FROM YEJIN: perhaps give more focus to the enhanced generation techniques in the intro}

%Overall, our contributions center around bringing the study of factual consistency to a new domain: \promptTask. We propose \emph{factual ablation}, a novel method for measuring factual consistency, and use this to generate evaluation data (both synthetic and natural). Finally, we propose multiple methods for improving factual consistency in this domain, carrying out a wide evaluation of models using lexical metrics, factual ablation, and human annotation. This present an exciting step for factual consistency outside of summarization.
Overall, our contributions bring the study of factual consistency to a new domain: \promptTask. We propose \emph{factual ablation}, then use this to generate evaluation data (both synthetic and natural). We propose multiple methods to improve factual consistency, carrying out a wide evaluation of models using lexical metrics, factual ablation, and human annotation, finding the superior model by factual ablation also achieves the best human-measured factual consistency. As natural generation models see increasing deployment, it is more important than ever to make sure they are factual and well controlled (\S\ref{sec:ethics}). Studying this in highly applicable domains, like content transfer, is an important step in keeping models accountable. 

%This present an exciting step for factual consistency 
%outside of summarization.
%in generation.

% next, talk about evaluation setups, then modeling, then overall conclusions etc

% talk about lms

% talk about grounded gen, like dialogue and wikipedia

% talk about other factuality concerns (summarization, a simpler setting)

% lay this out in steps: we first show data is ungrounded, then show this
% leads to a blind-spot in traditional evaluations (maybe fig 1)
% so we consider how this can be improved 

%% file: figs/fig_1.tex
\begin{figure}[t]
    \centering
    \includegraphics[width=0.9\linewidth]{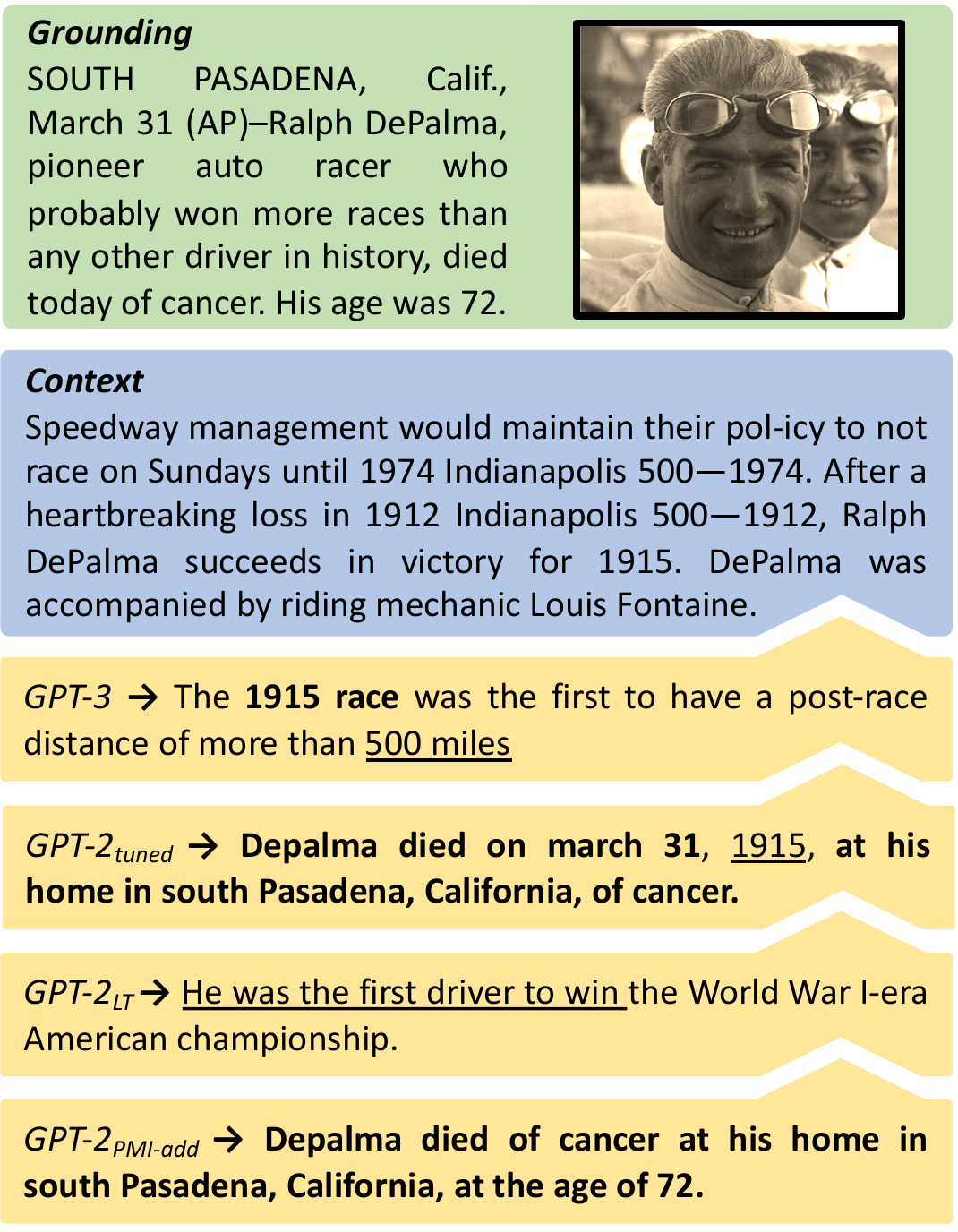}
    \caption{Generation with different models continuing a Wikipedia article. GPT-3 has no grounding, while the other 3 models
    use one document as grounding. The table highlights  \textbf{factual} and \underline{false} information. }
    \label{fig:fig_1}
\end{figure}

%% file: sections/background.tex
\section{Related Work and Background}
\label{sec:related}

\subsection{Textually Grounded Generation}

Textual grounding is a common element of natural language generation tasks, wherein a textual input is used to provide facts and information for decoding. One of the most popular tasks following this paradigm is abstractive summarization \citep{narayan2018don, rush2015neural}, in which generation $y$ should shorten and capture the salient information in source $g$. Other tasks extent beyond summarization, for example grounded dialogue \citep{Dziri2021EvaluatingGI} and \promptTask \citep{prabhumoye2019towards} (studied here). These tasks add the additional constraint that the generation $y$ must adhere to some existing context $c$, either previous dialogue turns or a document being extended (respectively).

\subsection{Factuality and Factual Consistency}
\label{subsection:background_factuality}

Recent work \cite{maynez-etal-2020-faithfulness} observes that strong neural models, although fluent and creative, often hallucinate information. Indeed, for all summarization models tested by \citet{maynez-etal-2020-faithfulness}, over 70\% of generations included information not directly entailed by the grounding $g$. However, they observe that some of this information is still factually correct. This naturally yields 2 notions of correctness for textually grounded generation: \textit{factuality} and \textit{factual consistency} (or \textit{faithfulness}). Factuality concerns the universal correctness of a generation--is the model output factual regardless of grounding $g$? Factual consistency more specifically probes whether the generation adheres to grounding $g$. Our work probes the much more tractable problem of factual consistency.

A significant portion of past work on factuality and factual consistency in generation has focused on abstractive summarization \cite{pagnoni-etal-2021-understanding,goyal-durrett-2021-annotating,cao-wang-2021-cliff,  aralikatte-etal-2021-focus}. Yet as mentioned above, textually grounded generation extends beyond summarization, and some works explore notions of factuality in other domains such as conversation \cite{shuster-etal-2021-retrieval-augmentation} or table-to-text generation \cite{ Liu2021TowardsFI}. Similarly, we explore these notions outside of direct summarization, instead focusing on grounded content transfer \cite{prabhumoye2019towards}.

Much work in this area concerns improving factuality and factual consistency \cite{shuster-etal-2021-retrieval-augmentation, zhu-etal-2021-enhancing, Nan2021ImprovingFC, Mao2020ConstrainedAS,aralikatte-etal-2021-focus}. While this is one aspect of our work, we also aim to improve automatic evaluation, for which a single standard metric has not emerged. Some works evaluate factuality and consistency with extraction \citep{goodrich2019assessing, zhang2020optimizing} or question answering \citep{wang2020asking, durmus-etal-2020-feqa, Nan2021ImprovingFC}. Others use notions of entailment \citep{falke-etal-2019-ranking}, or simply train end-to-end models to judge these aspects directly \citep{kryscinski-etal-2020-evaluating}. We instead focus on the effect of excluding relevant information from the grounding--for a factual model, removing this information should lower the probability of the ground-truth generation. 

Some works follow a similar intuition to ours. \citet{xie-etal-2021-factual-consistency} also understand factuality by estimating the effect of the source document on generative model output, although they explicitly mask relevant information while we offer a plausible alternative grounding. Similarly, \citet{xu-durrett-2021-dissecting} ablate information from a source document to understand aspects of conditional generation, although factuality is not the focus. 

Finally, some work in this area studies the need to evaluate metrics of factuality and consistency \citep{Gabriel2020GoFA, pagnoni-etal-2021-understanding}, and to generally characterize and annotate the mistakes of models \cite{maynez-etal-2020-faithfulness,pagnoni-etal-2021-understanding, goyal-durrett-2021-annotating}

%%%%%%%%%%%
%Usable textually-grounded models should adhere to some notion of factuality: if models may not tell the truth, they lack practical value. Thus, many recent works focus on evaluating factual aspects of models. Some attempt to define and compare facts in the generation and grounding using extraction \citep{goodrich2019assessing, zhang2020optimizing} or question answering \citep{wang2020asking, durmus-etal-2020-feqa}. Others use notions of entailment \citep{falke-etal-2019-ranking}, or simply train end-to-end models to judge these aspects directly \citep{kryscinski-etal-2020-evaluating}. Still other work conducts meta evaluation of factuality metrics \citep{Gabriel2020GoFA}.

%Our work is novel compared to these in two ways: measuring factual consistency in a domain \textit{distinct} from pure summarization, and introducing the \textit{factual ablation} measure, with accompanying evaluation datasets. 
%%%%%%%%

%\peter{Make sure you cover all of these}
%FactCC: \citep{kryscinski2019evaluating}

%\noindent
%asking questions: \citep{wang2020asking}

%\noindent
%on faithfulness: \cite{maynez2020faithfulness}

%\noindent
%go figure (Asli paper): \cite{Gabriel2020GoFA}

%\noindent
%feqa: \cite{durmus2020feqa}

%\noindent
%Assessing  the  factual  accuracy of generated text (Goodrich)

%\noindent
%Very related recent paper on hallucination
%\cite{Zhou2020DetectingHC}

\subsection{Loss Truncation}
\label{subsec:loss_truncation_backgroung}

Loss Truncation \citep{kang-hashimoto-2020-improved} improves conditional models by only training on the top-c examples, ranked by dynamically updated model loss. This is broadly applicable to conditional models with a noisy learning signal, and we include two baselines using this approach.

%% file: sections/methodology.tex
\section{Methodology}

Here, we bring factual consistency to a new domain, \promptTask, which is the task of extending context $c$ with content from a grounding document $g$. We discuss the task (\S\ref{subsec:task}), and our major contributions: novel methods for judging (\S\ref{subsec:methodology_metric}) and improving (\S\ref{subsec:modeling}) factual consistency in this setting. 

\subsection{Task: Content Transfer}
\label{subsec:task}

Recent work studying factual consistency has largely focused on summarization: models are given a source document $g$ (grounding) as input, and output a shorter summary text $y$ capturing the most salient information from $g$. Summarization is a natural domain to study factual consistency--the source document typically contains all information needed for the summary--but the need for factual consistency is not exclusive to summarization, and more domains should be explored.

Here, we expand this study to the \textit{\promptTask} task. As in summarization, models are given grounding $g$, and must output text $y$ using information from $g$. However, $y$ must also fit a context $c$,
which significantly narrows the range of reasonable outputs from the open-ended summarization task, to those that fit the context. 
%In effect, factual consistency here does not require considering as wide a range of information from $g$. 
\citet{prabhumoye2019towards} also note the ineffectiveness of extractive methods for this task. This obviates issues of model understanding that underlie factual consistency errors: while summarization models can often copy text directly, ensuring factual consistency regardless of understanding, \promptTask models \emph{must} reformulate information to fit the context. 

\citet{prabhumoye2019towards} introduces this task, and we follow their use of Wikipedia data for content transfer: given a partial Wikipedia article $c$, models extend $c$ with a next-sentence $\hat{y}$, using information from the grounding document $g$ referenced by the true next-sentence $y$; $g$ contains the factual basis for $y$. %Quality of grounding documents $g$ was ensured by filtering to only include web pages from a set of trusted news sites, and further processing for quality.
The dataset contains 600K training examples, 6K validation examples, and 50K test examples. Measuring factual ablation on this original dataset is not an option as there is only one piece of grounding per-example, and so we describe two paths to generating evaluation data for this purpose below.

Content transfer is formally defined as the task of generating a next-sentence $\hat{y}$ for context $c$ which is (i) coherent, and fits $c$ (ii) factually and (iii) stylistically, while (iv) only utilizing information from grounding document $g$. Note here, (iv) requires factual consistency, which is a stronger notion than overall factuality (\S\ref{subsection:background_factuality}): We don't allow models to introduce facts that are not directly entailed by $g$. Even strong pretrained models can make factual errors when writing from memory (Figure~\ref{fig:fig_1}). 

Central to our study is the degree to which each above condition must be met to have an effective model. Conditions i-iii are not absolute constraints. A reasonable generation may be a bit awkward or not perfectly fit $c$. On the other hand, an effective model \emph{must} follow condition iv completely. While satisfaction of all of i-iv may be noisy in both the training dataset and tuned models, our approach will focus on addressing this noise for condition iv.

\subsection{Measure: \measure}
\label{subsec:methodology_metric}

Although the \promptTask dataset from \citet{prabhumoye2019towards} includes evaluation data, it takes a standard reference-comparison format, wherein a ground-truth target $y$ is provided for comparison with generations. Automatic comparison between generations and a reference does not specifically test for factual consistency; indeed lexical overlap metrics show low correlation with notions of factuality (e.g. ROUGE in \citealt{falke-etal-2019-ranking}). Thus, we propose a new measure--\emph{factual ablation}--for judging factual consistency of models in this setting. To do this, we construct a secondary evaluation set.

Intuitively, \promptTask models should be less likely to output next-sentence $y$ as fewer facts in $y$ are supported by grounding $g$. Factual ablation tests this: As relevant facts are \textit{ablated} from $g$ ($\xrightarrow{}g'$) then $y$ should become less likely under a grounded generation model $P$, as it becomes less factually supported. To define this precisely, suppose we have 2 grounding documents $g$, $g'$ s.t. $g \implies y$ ($g$ \textit{entails} $y$) and $g' \centernot\implies y$, then we should have:

%To define this precisely, suppose function $F(X)$ returns the semantic facts contained in or clearly entailed by text $X$. Taking documents $g,g'$, factual ablation judges the degree to which the following holds:

\begin{equation} \label{eq:basic_acc}
    P(y|c,g) > P(y|c, g')
\end{equation}

\noindent
In words, model $P$ follows factual ablation if it prefers to generate target $y$ given grounding $g$ that entails $y$, over $g'$ that does not (i.e. contains a subset of the information necessary for $y)$. 

Factual ablation is a necessary condition for a completely factually consistent model\footnote{Given $P(y|c,g) >0$ for original grounding $g$}: if a model will only output facts contained in grounding $g$ (consistent), then $P(y|c,g') = 0$ as $g'$ contains only a subset of facts in $y$, by definition. As a proxy for factual consistency, factual ablation is also easier to measure directly. Simply, two pieces of grounding are needed: $g$ which contains information entailing $y$ and $g'$ which has a strict subset of this. Then we judge factual ablation for the model by comparing $P(y|c,g)$ and $P(y|c,g')$.

We propose a number of ways to compare these values. The most straightforward is \emph{accuracy}, the frequency of:
\begin{equation}
(accuracy) \text{   } P(y_i|c_i,g_i)>P(y_i|c_i,g'_i)
\end{equation}
%\begin{equation}
%acc = \frac{1}{N}\sum_{i=1}^N P(y|c,g)>P(y|c,g')
%\end{equation}
or how often model $P$ is less likely to produce target $y$ given ablated grounding $g'$. However, we are interested in the \textit{generative} qualities of the model $P$, whether having access to fewer relevant facts \textit{significantly} decreases generation probability for $y$. High accuracy only requires the probability drop, perhaps a trivially small amount, not indicative of the model's generation properties. Indeed, we find even a zero-shot language model (GPT-2) achieved accuracy close to tuned models (Table \ref{tab:cf_combined}). While the zero shot model detects changes in grounding, the difference is minute.

Thus, we offer a second metric that enforces a \textit{significant} change in probability-- \emph{margin-accuracy}, which is how often the following holds:
\begin{equation*} \label{eq:margin_accuracy}
    (acc_{marg}) \text{ } \log(P(y|c,g)) > m + \log(P(y|c,g'))
\end{equation*}
\noindent
where margin $m$ is a parameter. This comes with a simple interpretation: the number of examples where having less factual support \textit{significantly} decreases generation probability, with significance defined by margin $m$. For example, setting $m$ = $\text{log}(100)$ requires $y$ to be at least 100 times less likely under $g'$ than $g$ to be considered a success. 

In experiments, the margin giving the clearest spread of models is highly dataset-depending, with a smaller margin needed when grounding $g$ and ablated grounding $g'$ are more similar. The order of model performance will typically remain the same for different margins, but a poorly picked margin can result in less useful information--a large margin for datasets in which $g$ and $g'$ are close can result in most models close to 0 (too difficult) while a small margin when $g$ and $g'$ are far apart can similarly result in most models close to 100 (too easy). For example, taking $m=0$ corresponds to pure accuracy, which we find does not give much separation between model performance. We suggest picking a margin $m$ that results in an informative spread, or reporting multiple margins if this is difficult. 

While directly measuring factual consistency outside of human evaluation is complicated, factual ablation is easily measured by constructing datasets with grounding pairs $g,g'$. We construct both a handcrafted synthetic set with manually ablated grounding (\S\ref{subsec:data_synthetic}) and a natural set which leverages the edit structure of Wikipedia (\S\ref{subsec:data_wiki}). Note that grounding $g,g'$ should be as similar as possible while still correct, for a meaningful and challenging example.

\subsubsection{Synthetic Evaluation}
\label{subsec:data_synthetic}

Deliberate and purposeful edits offer a simple path to evaluating aspects of models \citep{Ribeiro2020BeyondAB}. As such, one approach we offer for generating evaluation data for factual ablation is using handcrafted examples, by editing. We make point-edits to the grounding document $g$ to produce $g'$ which has strictly fewer facts in common with target $y$, easily producing correct and interpretable factual ablation examples. 

We construct a set of synthetic examples by editing single pieces of information in both the grounding $g$ and target $y$, producing $g'$ and $y'$ which share this modified fact. This yields two examples:

$$(g,g',c,y)\text{ and } (g',g,c,y')$$
\noindent
where $y$ should prefer $g$ and $y'$ should prefer $g'$. We limit edits to two types of information: numerical (changing numbers: e.g., four miners became stuck $\xrightarrow{}$ two miners became stuck) and chronological (the Queen toured Canada in March $\xrightarrow{}$ the Queen toured Canada in April). These edits are only made for examples where (i) the fact is not commonly known (i.e. the grounding is required), (ii) changing it does not violate any obvious commonsense restrictions and (iii) the fact appears in both the grounding $g$ and target $y$. Our resulting dataset contains 162 such examples (see appendix for example). Note, from an ethical standpoint we avoid constructing examples related to sensitive topics or potential disinformation; synthetic factual ablation is useful at a small scale, but should not be done at a large scale for this reason.

While synthetic data is simple to produce and well-controlled, it has obvious drawbacks. Mainly, the style of factual differences produced will be limited and biased, and the number of examples relatively low as each must be handcrafted. To overcome these issues, we also introduce a natural evaluation set.

\subsubsection{Natural Evaluation}
\label{subsec:data_wiki}
The use of Wikipedia data for the original \promptTask dataset from \citet{prabhumoye2019towards} offers an intuitive way to construct natural evaluation data for factual ablation. Because Wikipedia is constantly edited, there are many instances where one sentence $y$ including a reference $g$, is replaced by another pair $y'$, $g'$. In practice, $y,y'$ will tend to be \textit{entailed} by their own grounding ($g,g'$ respectively) and not the other. This means $g$ can serve as ablated grounding for $y'$ and vice versa. We are also ensured that both $g,g'$ can result in a reasonable continuation to $c$, which ensures that examples are not trivial. Selecting such a document automatically would be challenging: if it is too unrelated the example it becomes trivial, while a relevant document may not be considered ablated at all (i.e. it may contain as much relevant information as the original). 
The Wikipedia-edit dataset is constructed as follows:

\begin{enumerate}[nosep]
    \item Isolate all instances $(g,g',c,y,y')$ in Wikipedia edit data where referenced sentence $y$ has been replaced by referenced sentence $y'$.
    \item From each such instance, construct two \measure examples: $(g,g',c,y)$ and $(g',g,c,y')$.
    \item Filter any such examples that do not meet quality criteria.
\end{enumerate}
\noindent
We impose a number of quality criteria on examples $(g,g', c,y)$, imposing $y$ is between 50 and 200 character, $c$ up to 3 sentences, $g$ and $g'$ come from news sites and can be fully recovered, no text includes excessive formatting issues. We will release processing code with the dataset. We attempt to recreate a similar distribution to the \promptTask dataset of \citet{prabhumoye2019towards}, following the same post processing steps. This prevents major domain transfer issues between our training and testing. In total, we extract 710 examples, although larger sets can be constructed as Wikipedia is constantly being edited. See appendix for a full example. 

\subsection{Modeling}
\label{subsec:modeling}

Models tuned directly on grounded generation data often violate factual consistency. In \citet{maynez-etal-2020-faithfulness}, over 70\% of generated summaries were found to contain factual inconsistencies with respect to the grounding, and in our own experiments a model tuned on \promptTask data has similar shortcomings ($\text{GPT-2}_{tuned}$ in Figure~\ref{fig:fig_1}).

Yet these models often generate \textit{some} factually correct information. Clearly a notion of factual consistency is being modelled, but this is not represented strongly enough at generation time. We consider two approaches to rectify this: removing data points that may be encouraging inconsistency at training time (\S\ref{subsubsec:training_time_methods}), and inflating this consistency signal at inference time (\S\ref{subsubsec:inference_time_methods}).

\subsubsection{Training-Time Methods}
\label{subsubsec:training_time_methods}
Loss truncation \citep{kang-hashimoto-2020-improved} is a training technique that works by only training on the top-c fraction of examples by loss, calculated dynamically as training proceeds. This follows the intuition that \textit{degenerate} training examples which erode model performance will be difficult to predict even as training progresses, and can thus be selected out. In our case, this corresponds especially to examples where target $y$ contains facts outside of grounding $g$, limiting predictability. We test this original form of loss truncation, with parameter $c$ indicating the degree of examples to ignore ($1-c$).

Loss Truncation is general to many tasks, but does not consider specific signals in grounded generation. We extend the method to take this into account, in a ``grounded'' version. Here, we additionally truncate $1-c_{gnd}$ of training examples, by the amount \textit{grounding improves loss}, given by:
\begin{equation}
\label{eq:gnd_gap}
    \log P(y|c,g) - \log P(y|c)
\end{equation}
where $P(y|c,g)$ is estimated by the training model, and $P(y|c)$ by a model tuned to predict y based only on c (ungrounded). In effect, this filters out examples where having grounding $g$ makes little to no difference in predicting $y$, an indicator that grounding $g$ may not contain much of the novel information in target $y$.

\subsubsection{Inference-Time Methods}
\label{subsubsec:inference_time_methods}
Following a similar intuition to grounded loss truncation (above), we propose algorithms to improve factual support at inference time. At training time, we use the amount that grounding $g$ improves prediction probability (equation \ref{eq:gnd_gap}) as a signal for which targets $y$ actually use information from $g$. We hypothesize that we can make more use of grounding at \textit{inference time} by following this same signal of how much text probability increases with grounding $g$. Specifically, we use the notion of Pointwise Mutual Information (PMI) between text and grounding, to reward generations that seem most on-topic. We propose and test multiple ways this can be realized:

\paragraph{PMI-Interpolation} specifically estimates how well supported text is by grounding using (PMI), holding context $c$ constant:
\begin{equation}
    s_{pmi}(t_i;g) = \text{log}\frac{P(t_i|g, c, t_{0:i-1})}{P(t_i| c, t_{0:i-1})}
\end{equation}
PMI-Interpolation is defined in the log-scale, by interpolating $s_{pmi}$ with the logits of $P(t_i|g, c, t_{0:i-1})$, then taking a softmax to define full probability, i.e.
\begin{align}
    P_{pmi-interp} \propto \exp\big(& (1 - \alpha)\log P(t_i|g, c, t_{0:i-1}) \nonumber \\ & + \alpha s_{pmi}(t_i;g)\big)
\end{align}
where $\alpha \in [0,1]$ is a mixing parameter controlling the effect size of $s_{pmi}$. $\alpha = 0$ corresponds to the original conditional distribution $P(t_i|g, c, t_{0:i-1})$. This method is equivalent to taking a Product of Experts \citep{hinton2002training} between $P(t_i|g, c, t_{0:i-1})$ and a softmax distribution  of PMI between each token and the grounding. 

\paragraph{PMI-Addition} follows a similar intuition to PMI-Interpolation. Rather than mixing $P(t_i|g, c, t_{0:i-1})$ with a distribution defined by PMI, we add $s_{pmi}$, rewarding tokens which are estimated to share information with the grounding:
\begin{align}
    %P&_{pmi-add} \propto \nonumber \\ &e^{\text{log}(P(t_i|g, c, t_{0:i-1})) + \alpha*s_{pmi}(t_i;g)}
    P_{pmi-add} \propto \exp\big( & \log P(t_i|g, c, t_{0:i-1})\nonumber \\ & + \alpha s_{pmi}(t_i;g) \big)
\end{align}
$\alpha \in [0,1]$ controls how much we reward tokens with high PMI, up to adding the full PMI to the generation model's logits. 

%% file: sections/experiments.tex
\section{Experimental Setup}
\label{sec:experiments}

\input{figs/fig_lexical}

We probe factual consistency for an array of models tuned on the training set for content transfer from \citet{prabhumoye2019towards} (\S\ref{subsec:task}). We generate on the validation set, assessing the generations of each model with lexical and human metrics; then, we compare generative properties to the factual ablation of each model, measured on our synthetic (\S\ref{subsec:data_synthetic}) and natural (\S\ref{subsec:data_wiki}) evaluation sets. 

\subsection{Models}

All models tuned here follow the GPT-2 (small) architecture \citep{radford2019language}. We use the Huggingface \citep{Wolf2019HuggingFacesTS} library, with default parameters for training. We elaborate below.

\input{figs/fig_cf_combined}

\paragraph{Untuned}
We include some models that are not tuned on the \promptTask dataset (\S\ref{subsec:task}), but can be seen as transfer or zero-shot models. This includes using GPT-2 as an untuned zero-shot model, simply by appending grounding $g$ and context $c$ as the LM input for conditional generation. 

%Similarly, we treat the T5 seq2seq pretrained model as a zero-shot baseline. One task it is trained for is summarization by feeding a document into the encoder (prepended with ``summarize: ''), then allowing the decoder to generate a summary. To apply it to \promptTask we treat g as the input document, and take $c$ as a prefix to generation for the decoder (i.e. treating $c$ as a portion of the summary that has already been written). 

We also investigate how a model trained to judge factual consistency performs on the factual ablation task. We use the BERT-based \citep{Devlin2019BERTPO} FactCC model \citep{kryscinski-etal-2020-evaluating}, which is trained to judge the factual consistency between a document and summary. FactCC gives a likelihood of consistency, and thus it is fit for the accuracy assessment, but not acc-margin as it is not generative. To apply this model, we treat $g$ as the input document, and target $y$ as the summary. Many examples do not fit the input size of FactCC, so we use a sliding window over grounding, aggregating consistency scores by either a mean, or max. 

\paragraph{Tuned}
We include 2 basic finetuned models. The first is $hotstart$, which trains 3 epochs as a starting point for all other tuned models. Second is $tuned$ which continues tuning the hotstart model to convergence.

\paragraph{Loss Truncation}
As discussed in \S\ref{subsec:modeling}, we consider 2 forms of loss truncation: basic and ``grounding'' , denoted here by $LT_{basic}$ and $LT_{+gnd}$. Both of these begin with the hotstart model, but apply loss truncation as discussed in \S\ref{subsec:modeling}, with parameter $keepc=0.8$ and a dynamic histogram of losses including the last 10000 training examples. 

\paragraph{Inference-Time}
Finally, we test both inference-time algorithms from \S\ref{subsec:modeling}. Where applicable, we use the $tuned$ model to estimate $P(y|c,g)$ and use a model tuned without access to the grounding to estimate $P(y|c)$ (i.e. in each training example, g is replaced by the empty string). 
\emph{PMI-Interpolation} models are denoted $PMI_{interp}$ and we consider $\alpha$ values of $0.1,0.3,0.5$.
\emph{PMI-Addition} models are denoted $PMI_{add}$ and we consider $\alpha$ values of $0.1,0.3,0.5$.

\subsection{Experiments}

\subsubsection{Content Transfer Generation}
In this experiment, we explicitly test the generative qualities of each model by generating content transfer document completions on the validation set from \citet{prabhumoye2019towards}. Models generate using top-p sampling \citep{holtzman2019curious} with $p=0.5$, until 1 full sentence is produced. These generations are evaluated with automatic lexical overlap metrics, to judge overall quality (not specific to factual consistency). We also carry out a pairwise human evaluation on these. We include generation examples in the appendix.

\paragraph{Data} We generate with each model on the 6K examples in the content transfer validation set  (\S\ref{subsec:task}).

\paragraph{Metrics}

We use a set of automatic lexical metrics, as in  \citet{prabhumoye2019towards}. We measure NIST \citep{NIST}, BLEU \citep{papineni2002bleu}, and METEOR \citep{denkowski2014meteor} as a cross-section of common metrics. As discussed in \S\ref{subsec:methodology_metric}, lexical metrics do not give a strong signal for factual consistency, but can help understand the tradeoff between this and other notions of quality (conditions i-iii from \S\ref{subsec:task}). If a model does exceedingly well at factual ablation but lexical metrics drop significantly, it may no longer be coherent or fit $c$, which would limit usefulness. 

Further, we carry out a small-scale human evaluation on these generations, asking about (i) fluency and fit with context $c$ and (ii) factual consistency, as the degree to which the generation $\hat{y}$ is supported by the grounding. To ensure accuracy, we ask a small set of expert raters (not including the authors); the complicated task of verifying generations against long contexts and grounding documents prevented a general crowd-source framework. We select for relatively short grounding documents (up to 300 words) and carry out a pairwise comparison between an inference-time algorithm that has a good balance of lexical and factual ablation scores ($PMI_{add,\alpha=0.3}$) and 2 baselines: the vanilla \emph{tuned} model and $LT_{basic}$. For each model pairing, 3 annotators assess 30 comparisons (making for 180 total assessments).
%We find Krippendorff's $alpha$ (ordinal) of 0.315 (fluency/context) and 0.496 (factual support). 
%% Sorry I independently wrote my own version elsewhere. Feel free to pick whichever version you prefer.
We used ordinal Krippendorff’s alpha \cite{krippendorff2007computing} for measuring inter-annotator agreement
%, as it accounts for ordinal  judgments.
which yields a coefficient of .331 for fluency and .393 for factual support. This is on a range from -1, to 1, and both values are considered ``fair''. The results of this study are included in Table~\ref{tab:human}. %The latter agreement is quite encouraging considering the amount of reading required from the judges.

\subsubsection{Testing \measure}

Here, we explicitly measure factual ablation across tested models using our constructed evaluation sets.  

\paragraph{Data} We carry out a factual ablation evaluation on our 2 generated datasets. Our synthetic dataset (\S\ref{subsec:data_synthetic}) contains 162 handcrafted examples, created by manually ablating facts from examples in the evaluation set from \S\ref{subsec:task}. Our natural dataset (\S\ref{subsec:data_wiki}) contains 710 examples, and is constructed by isolating instances where Wikipedia is edited to replace one grounded sentence $y$ with another $y'$ that uses different grounding. 

\paragraph{Metrics} We apply the accuracy and margin-accuracy metrics defined in \S\ref{subsec:methodology_metric}. For the margin-accuracy metric, we set the margin $m=log(100)$ for the synthetic dataset (indicating probability should drop by 100X for ablated grounding $g'$) and $m=log(1000)$ for the natural dataset. 
%Generally, we found the margin giving the clearest spread of models is highly dataset-depending, with a smaller margin needed when grounding $g$ and ablated grounding $g'$ are more similar; we use a smaller margin for the synthetic dataset as the ablation is made by point-edit, and is likely to cause a much smaller factual difference between $g$ and $g'$. The order of model performance will typically remain the same for different margins, but a poorly picked margin can result in less useful information--a large margin when $g$ and $g'$ are close can result in most models close to 0 (too easy) while a small margin when $g$ and $g'$ are far apart can similarly result in most models close to 100 (too difficult). We suggest picking a margin $m$ that results in an informative spread, or reporting multiple values if this is difficult. 

\input{figs/fig_human}

%% file: figs/fig_lexical.tex
\begin{table}[t]
\small
    \centering
    \begin{tabular}{l|ccc}
     \toprule
       & NIST & BLEU & METEOR \\
   \midrule
   \rowcolor[gray]{0.95}
   \multicolumn{4}{l}{\emph{Tuned}} \\
   $hotstart$ & \textbf{2.0} & 11.3 & 6.8 \\
   $tuned$ & 1.8 & 11.9 & 7.3\\
    \midrule
    \rowcolor[gray]{0.95}
    \multicolumn{4}{l}{\emph{Loss Truncation}} \\
   $LT_{basic}$ & 1.8 & \textbf{12.1} & \textbf{7.4}\\
   $LT_{+gnd}$ & 1.8 & 12.0 & \textbf{7.4} \\
   \midrule
   \rowcolor[gray]{0.95}
   \multicolumn{4}{l}{\emph{Inference-time}}\\
   $PMI_{interp, \alpha=0.1}$ & 1.5 & 10.9 & 7.1\\
   $PMI_{interp, \alpha=0.3}$ & 1.6 & 9.7 & 6.4\\
   $PMI_{interp, \alpha=0.5}$ & 1.0 & 4.5 & 3.5\\
   \midrule
   $PMI_{add, \alpha=0.1}$ & 1.4 & 11.0 & 7.2\\
   $PMI_{add, \alpha=0.3}$ & 1.4 & 10.9 & 7.3\\
   $PMI_{add, \alpha=0.5}$ & 1.4 & 10.6 & 7.1 \\
     \bottomrule
    \end{tabular}
    \caption{Lexical generation evaluation on the validation set for \promptTask from \citet{prabhumoye2019towards}.}
     
    \label{tab:lexical}
\end{table}

%% file: figs/fig_cf_combined.tex
\begin{table}[t]
\small
    \centering
    \begin{tabular}{l|cc|cc}
     \toprule
     \multicolumn{1}{c}{}& \multicolumn{2}{c}{Synthetic} & \multicolumn{2}{c}{Natural} \\
       & acc & acc$_{marg}$ & acc & acc$_{marg}$ \\ 
   \midrule
      \rowcolor[gray]{0.95}
   \multicolumn{5}{l}{\emph{Zero Shot and Transfer}}  \\
   FactCC (mean) & 70.1 & - &  30 & - \\
   FactCC (max) & 37.0 & - & 63.9 & - \\
   GPT-2-zs & 78.0 & 2.4 & 84.5 & 54.5\\
   \midrule
   \rowcolor[gray]{0.95}
   \multicolumn{5}{l}{\emph{Tuned}} \\
   $hotstart$ & 74.4 & 10.7 & 87.9 & 64.5   \\
   $tuned$ & 75.0 & 19.6 & 87.7 & 69.2\\
    \midrule
    \rowcolor[gray]{0.95}
    \multicolumn{5}{l}{\emph{Loss Truncation}} \\
   $LT_{basic}$ & 75.0 & 23.8 & 87.7 & 70.3 \\
   $LT_{+gnd}$ & 75.0 & 18.5 & 88.2 & 71.1 \\
   \midrule
   \rowcolor[gray]{0.95}
   \multicolumn{5}{l}{\emph{Inference-time}}\\
   $PMI_{interp,\alpha=0.1}$& 75.0 & 20.8 & 88.0 & 69.0  \\
   $PMI_{interp,\alpha=0.3}$& 75.0 & 21.4 & 88.6 & 71.3 \\
   $PMI_{interp,\alpha=0.5}$& 76.8 & 23.8 & 88.9 & 76.1  \\
   \midrule
   $PMI_{add,\alpha=0.1}$& 74.4 & 23.8 & 87.9 & 70.6 \\
   $PMI_{add,\alpha=0.3}$& 73.2 & 28.6 & 87.9 & 72.7 \\
   $PMI_{add,\alpha=0.5}$& 71.4 & 32.1 & 87.3 & 73.0  \\
     \bottomrule
    \end{tabular}
    \caption{Evaluation of factual ablation with accuracy and margin-accuracy. Left is our \emph{synthetic} dataset (\S\ref{subsec:data_synthetic}) based on manual edits to grounding and target, with margin of log(100). Right is our natural dataset (\S\ref{subsec:data_wiki}) based on Wikipedia edits, using a margin of log(1000).}  
    \label{tab:cf_combined}
\end{table}

%% file: figs/fig_human.tex
\begin{table}[t]
\small
    \centering
    \begin{tabular}{l|ccc|c}
     \toprule
       & fluency and context & factual support \\ 
       
    \midrule  
    $tuned$ & 47.2 & 59.4   \\
    $LT_{basic}$ & 50.6 & 61.7 \\ 
     \bottomrule
    \end{tabular}
    \caption{Pairwise evaluation between one of our models ($PMI_{add}$ with $\alpha=0.3$) and two baselines--the $tuned$ baseline and $LT_{basic}$. 50.0 Indicates a tie, while $>50$ indicates preference for our model.}  
    \label{tab:human}
\end{table}

%% file: sections/results.tex
\section{Results and Analysis}
%Broadly, our results show many paths to improving beyond a basic conditional baseline ($tuned$) using existing methods (e.g. $LT_{basic}$) and our own.

\paragraph{Lexical Overlap}
Lexical overlap metrics for model generations are reported in Table~\ref{tab:lexical}. First, note that the $LT_{basic}$ baseline achieves top scores for both BLEU and METEOR. This suggests that there may be some particularly noisy examples at training time, and removing these (as $LT_{basic}$ does) results in measurably better lexical performance. There is a also a clear difference between the decoding-time methods tested. While $PMI_{add}$ holds fairly consistent scores across tested $\alpha$ values, the scores of $PMI_{interp}$ drop quickly. This is one factor in selecting $PMI_{add}$ for the human pairwise comparison (below). Although \textit{high} lexical overlap does not ensure factual generations \citep{falke-etal-2019-ranking}, we found systems with very \textit{low} lexical scores were often too incoherent to be factual. 

\paragraph{Factual Ablation}
As mentioned in \S\ref{subsec:methodology_metric}, factual ablation accuracy scores fall within a very similar range across models, for both the synthetic and natural factual ablation studies (Table~\ref{tab:cf_combined}); the one exception is the low score of the out-of-domain factual consistency checker (FactCC). We focus on margin-accuracy (acc$_{marg}$) as it gives a better indication of differences in generation behavior. In both evaluation sets, $LT_{basic}$ does significantly better than $tuned$, while $LT_{+gnd}$ does not have consistent performance across the sets. $PMI_{interp}$ and $PMI_{add}$ both show increasingly large advantages over other models as $\alpha$ is increased. However, the unstable performance of $PMI_{interp}$ on lexical metrics motivates choosing $PMI_{add}$ for our pairwise human evaluations, setting $\alpha=0.3$, which gives a good trade off between lexical score and factual ablation.

\paragraph{Human Evaluation}
Table~\ref{tab:human} compares $PMI_{add,\alpha=0.3}$ to the basic $tuned$ baseline and loss truncation $LT_{basic}$ \cite{kang-hashimoto-2020-improved}. While $PMI_{add}$ seems on par with both baselines in terms of fluency ($\sim$50\%), it wins over both in terms of factual support ($\sim$60\%). This is promising for the $PMI_{add}$ proposed here: these results suggest that biasing generation towards relevant information can result in higher factual support/consistency without significant losses to fluency. Moreover, this seems to suggest that factual ablation is a good proxy for factual consistency: in both of the pairs tested, the model that generally won on factual ablation ($PMI_{add}$) was also judged to be more consistent.

\paragraph{Discussion and Future Work}
In the future, inference-time strategies may be improved by using a lower noise (higher quality) estimator like $LT_{basic}$ rather than the basic conditional $tuned$ model. We avoid this for the sake of fair comparison between baselines. Second, it will likely be advantageous to add an explicit measure for fluency or linguistic smoothness when evaluating inference-time methods in particular, which risk disfluency. Clearly it is possible to go overboard (e.g. for $PMI_{interp,\alpha =0.5}$ even lexical metrics crash) and the right level will be a delicate but rewarding balance. This shouldn't discourage inference-time methods. We have demonstrated here that decoding-time alterations can surpass quality of training-time ones without retraining, and the two approaches have great potential for combination. Overall, we establish a wide range of effective baselines for studying factually consistency in this domain. (see \S\ref{app:gen_examples} for generations)

The agreement between human evaluation and factual ablation in this setting is a promising sign of the usefulness of this measure. Further, unlike model-based methods for measuring factuality and consistency \cite{wang2020asking, kryscinski-etal-2020-evaluating}, factual ablation is not limited by the quality of existing models--rather, the quality of the measure is linked to the quality of its evaluation set which can be validated and expanded by humans.
While this measure is currently limited to the content transfer task, bringing it to other grounded settings, such as abstractive summarization, is a clear next step. 

%% file: sections/conclusion.tex
\section{Conclusions}
In this work, we introduce the study of factual consistency to the content transfer domain by proposing factual ablation, a measure of factual consistency that uniquely fits this setup. We test multiple training-time and inference-time methods for improving factual consistency in this domain, carrying out a wide study of lexical metrics, factual ablation, and pairwise human comparison. We find the same model is superior at both factual ablation and human-judged factual consistency; this supports factual ablation as a useful measure in developing more consistent models, extending the already rich and promising vein of methods studied here.

\section{Ethical Considerations}
\label{sec:ethics}

We believe that work on grounded generation models and specifically on probing factual consistency in such models has positive implications for Ethics in AI, especially in the terms of addressing the potential harms and misuses \cite{bender:21} of large pre-trained models such as GPT-3 \cite{Brown2020LanguageMA}. Bender et al. have shown that such large pre-trained models can easily be led to generate inaccurate, offensive, and otherwise harmful texts. Such pitfalls motivate making text generation more controllable and {\it grounded}, as grounding amounts to constraining where semantic content originates, and this can help prevent the use of erroneous or outdated information. But even grounded generation is sometimes prone to generating factually incorrect texts, and our work helps fulfill the need to probe and increase the level of {\it factual consistency} between generated texts and trusted information sources.   

In terms of potential misuses of our work, we believe it is mostly tied to the users being potentially ill intended. While most users would probably make ethical use of controllable and grounded generation, we cannot completely ignore the risk of some users wanting to control generation to produce, e.g., fake news from dubious information sources (However, in this case we would argue it is mostly the user rather that AI that is at fault.) Nevertheless, the broader agenda of this work on factual consistency checking could also be helpful, as such dubious sources would contradict fact-checked information sources.

Regarding our handling of data and human subjects: Our work introduces two new evaluation datasets (\S\ref{subsec:data_synthetic},\ref{subsec:data_wiki}). Both are constructed using publicly accessible Wikipedia data only. Any modifications to this data (\S\ref{subsec:data_synthetic}) are made by authors of this paper only (i.e., no crowd-source human annotation). 
We also conducted a human evaluation that was small-scale on a volunteer basis by colleagues of the authors, and thus wide-scale payment is not a concern. Evaluation uses a simple multiple-choice input form, which offers no avenue for privacy concerns. 

% Highly recommended according to ACL instructions. Can be on the 9th page.

%% file: sections/appendix.tex
\section{Appendix}

\subsection{Factual Ablation Examples}
\label{app:fa_examples}

We include an example from the natural factual ablation dataset \S\ref{subsec:data_wiki} in Figure ~\ref{fig:ex_wiki}. We include an example from the synthetic factual ablation dataset \S\ref{subsec:data_synthetic} in Figure~\ref{fig:ex_synth}.

\subsection{Generation Examples}
\label{app:gen_examples}

We demonstrate generations for all models on an example from the content transfer dataset \S\ref{subsec:task}. See Figure~\ref{appendix_gen_examples}

\subsection{Human Evaluation}
\label{app:human_eval}

Here, we include the template used for pairwise human evaluation: Figure~\ref{fig:template}.

\input{figs/appendix_examples_wiki}
\input{figs/appendix_example_synth}

\input{figs/appendix_example_Gen}

\input{figs/template}

%% file: figs/appendix_examples_wiki.tex
\begin{figure*}[t]
    \centering
    \includegraphics[width=0.9\linewidth]{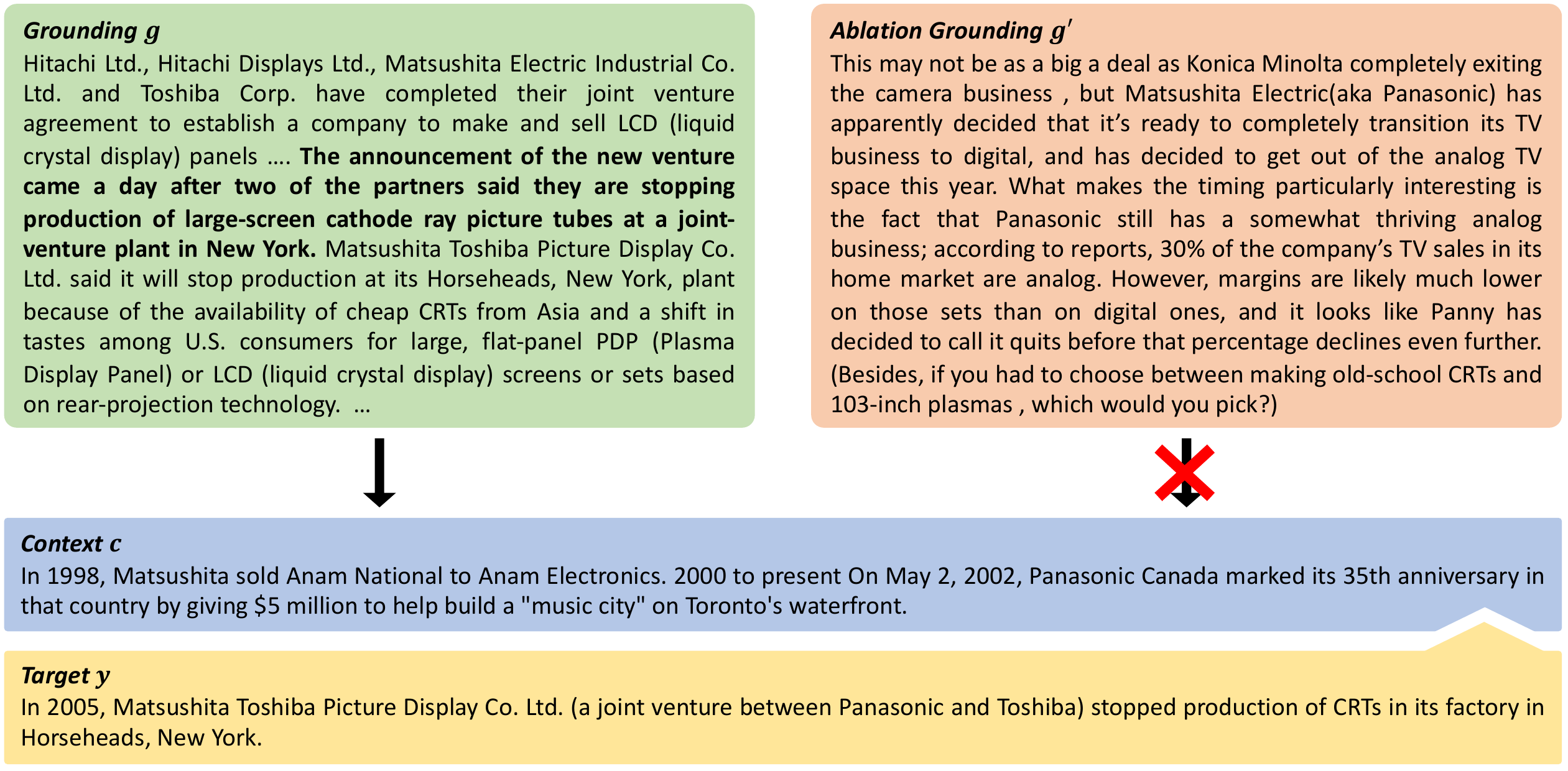}
    \caption{An example from the natural factual ablation dataset of \S\ref{subsec:data_wiki}. Relevant information is \textbf{bolded}. Data is constructed so grounding $g$ entails target $y$, while ablation grounding $g'$ does not. }
    \label{fig:ex_wiki}
\end{figure*} 

%% file: figs/appendix_example_synth.tex
\begin{figure*}[t]
    \centering
    \includegraphics[width=0.9\linewidth]{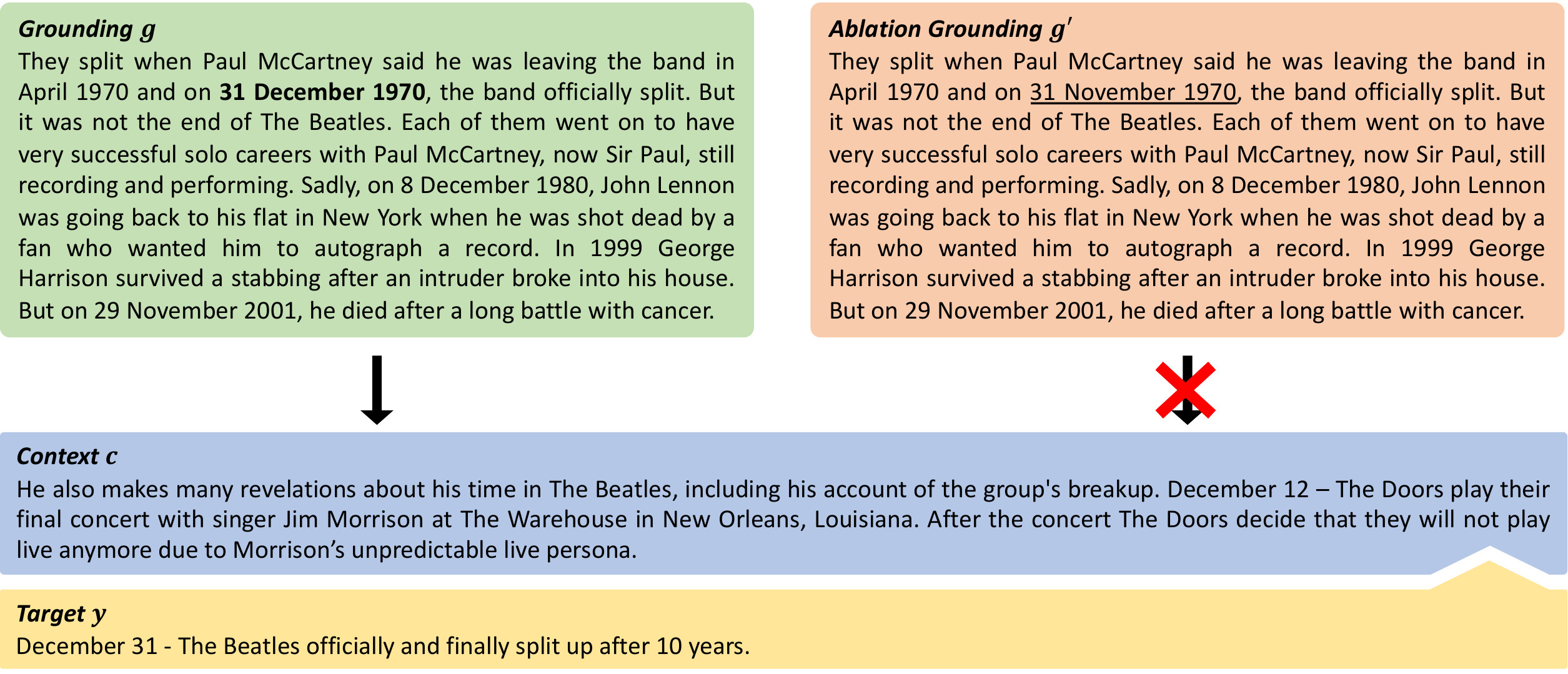}
    \caption{An example from the synthetic factual ablation dataset of \S\ref{subsec:data_synthetic}. Relevant information is \textbf{bolded}, and altered (ablated) information is \underline{underlined}. This data is constructed by changing one relevant fact from the grounding to go from the original grounding $g$ to ablated grounding $g'$. }
    \label{fig:ex_synth}
\end{figure*}

%% file: figs/appendix_example_Gen.tex
\begin{figure*}[t]
    \centering
    \includegraphics[width=0.9\linewidth]{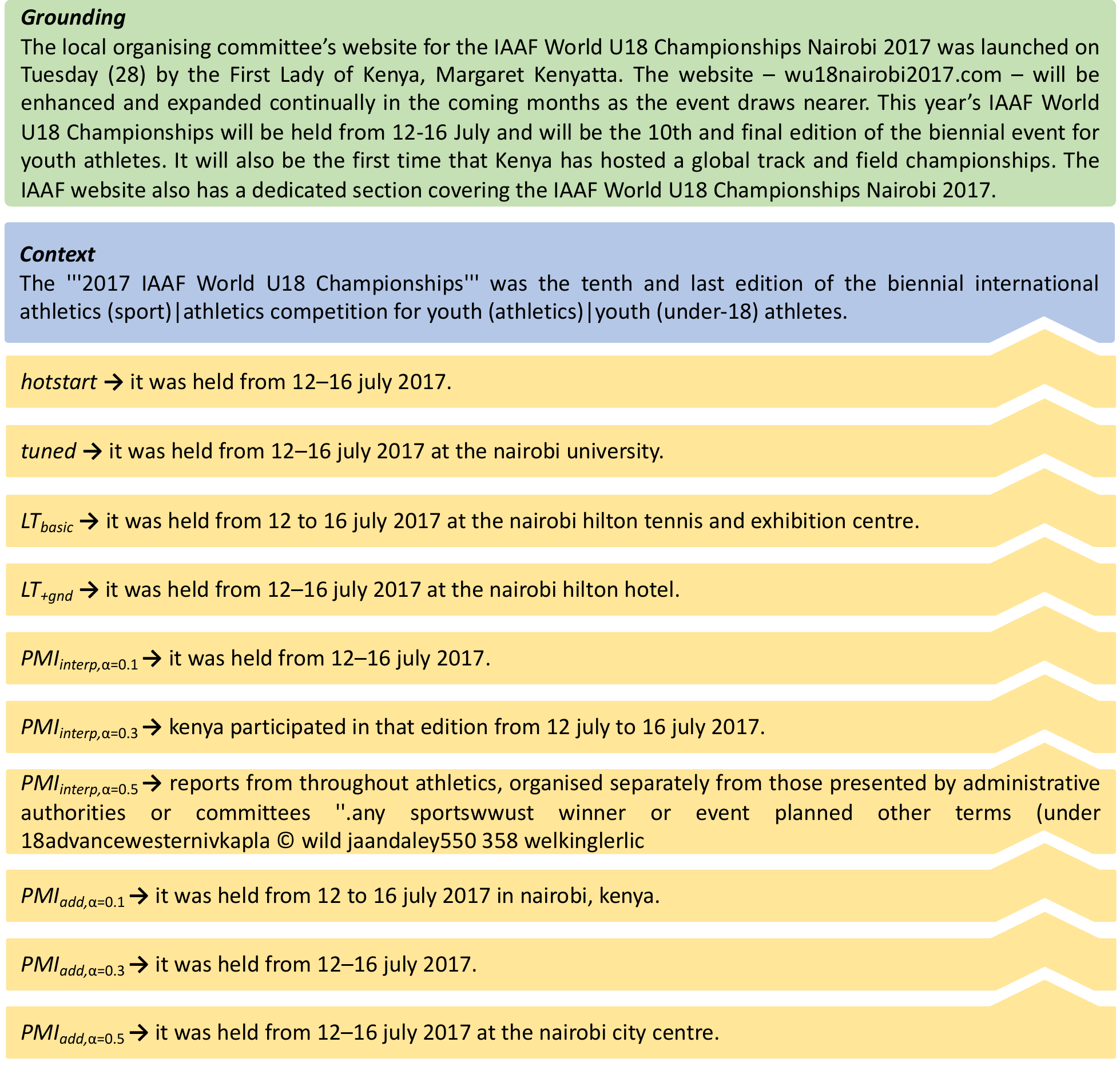}
    \caption{Example generations from all models tested. Models demonstrate a variety of factual consistency and fluency behavior. }
    \label{appendix_gen_examples}
\end{figure*} 

%% file: figs/template.tex
\begin{figure*}[t]
    \centering
    \includegraphics[width=0.9\linewidth]{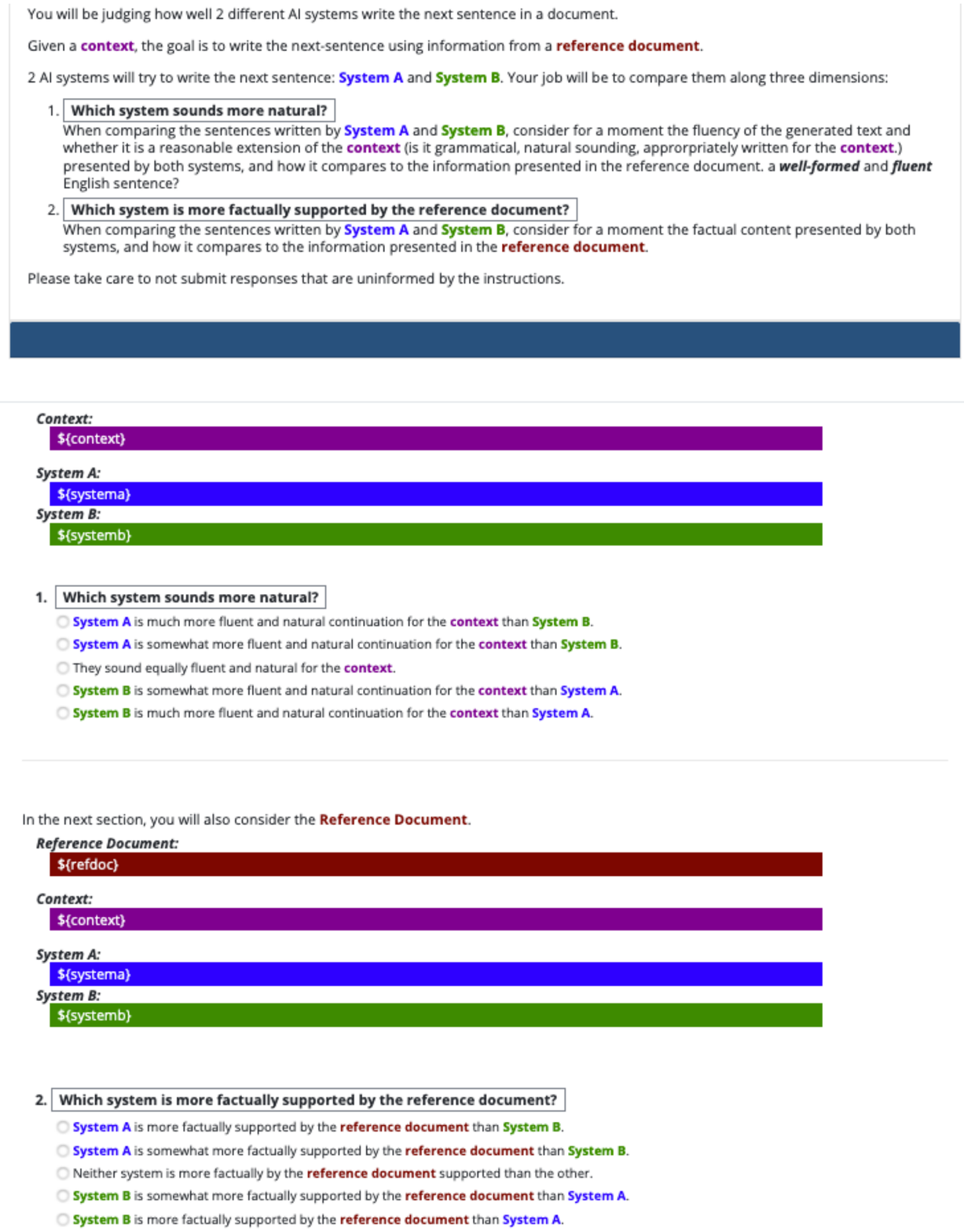}
    \caption{The template used for pairwise human evaluation
    }
    \label{fig:template}
\end{figure*}